\documentclass[11pt]{article}

\usepackage{acl}

\usepackage{times}
\usepackage{latexsym}
\usepackage[T1]{fontenc}
\usepackage[utf8]{inputenc}
\usepackage{microtype}
\usepackage{inconsolata}
\usepackage{graphicx}

\usepackage{amsmath,amssymb}
\usepackage{booktabs}
\usepackage{array}
\usepackage{enumitem}
\usepackage{tikz}
\usepackage{array}

\newcolumntype{R}[1]{>{\raggedleft\arraybackslash}p{#1}}
\newcolumntype{C}[1]{>{\centering\arraybackslash}p{#1}}
\usetikzlibrary{arrows.meta,positioning}

\usepackage[most]{tcolorbox}
\definecolor{lmadblue}{HTML}{0072B2}
\definecolor{lmadgreen}{HTML}{009E73}
\definecolor{lmadorange}{HTML}{D55E00}
\definecolor{lmadgold}{HTML}{E69F00}
\definecolor{lmadgray}{HTML}{7B8794}
\newtcolorbox{casetrace}[1]{
  enhanced,
  colback=white,colframe=lmadgray!75,
  boxrule=0.6pt,arc=2.5pt,
  left=7pt,right=7pt,top=6pt,bottom=6pt,
  fontupper=\footnotesize,
  title={#1},
  fonttitle=\bfseries\footnotesize\sffamily,
  coltitle=white,colbacktitle=lmadblue!90,
}
\newcommand{\casestage}[2]{\par\vspace{6pt}{\noindent\sffamily\bfseries\footnotesize\textcolor{#1}{#2}}\par\vspace{2pt}}
\newcommand{\nodetag}[1]{{\scriptsize[\textsc{#1}]}}
\newcommand{\okmark}{\textcolor{lmadgreen}{$\checkmark$}}
\newcommand{\errmark}{\textcolor{lmadorange}{$\times$}}
\newcommand{\commitclaim}[1]{\noindent\colorbox{lmadgreen!13}{\parbox{\dimexpr\linewidth-2\fboxsep\relax}{#1}}}

\setlist[itemize]{leftmargin=*,nosep}
\setlist[enumerate]{leftmargin=*,nosep}

\definecolor{commentblue}{RGB}{0,0,180}
\definecolor{commentgreen}{RGB}{0,180,180}

\newcommand{\method}{\textsc{LMAD}}

\title{Where Reasoning Diverges: \\
Localized Multi-Agent Debate for Multi-Hop Question Answering}
\author{
Weijun Gao\textsuperscript{1},
Xiang Ding\textsuperscript{2},
Haoyang Liu\textsuperscript{3},
Tiancheng Xing\textsuperscript{4}
\\[4pt]
\textsuperscript{1}The Chinese University of Hong Kong
\\
\textsuperscript{2}Nagoya University
\\
\textsuperscript{3}University of Illinois Urbana-Champaign
\\
\textsuperscript{4}Institute of Science Tokyo
}

\begin{document}
\maketitle

\begin{abstract}
Multi-agent debate commonly exchanges complete rationales even when disagreements concern only a few intermediate claims. We introduce \emph{Localized Multi-Agent Debate} (\method{}), an inference-time protocol that represents agent rationales as nodes, locates their earliest conflict, and restricts debate to the corresponding local segments. Guarded resolution extends a shared committed state so that later conflicts can be addressed without reopening accepted steps. We evaluate \method{} on four multi-hop question-answering benchmarks using ten backbones from four model families. Our method achieves the highest macro-averaged judge accuracy across all ten backbones, outperforming the strongest conventional baseline by up to 7.20 percentage points.
\end{abstract}

\section{Introduction}

Multi-agent debate (MAD) improves language-model reasoning by letting several model instances propose, inspect, and revise solutions before a final decision~\citep{pmlr-v235-du24e,chen-etal-2024-reconcile}. Existing protocols differ in how they sustain disagreement~\citep{cui-etal-2026-free} or reach a final decision~\citep{liu-etal-2026-ems}, but their shared state is usually an entire answer or rationale. Consequently, an agent that changes its answer must reconsider a long rationale containing both useful agreements and a small number of decisive errors.

This granularity mismatch is especially consequential in multi-hop question answering, where each intermediate claim constrains the next inference. Agents may agree on early facts but select different entities or relations at a later hop. That first disagreement propagates through the remaining steps, eventually producing different answers. Answer-level voting hides where the divergence began. The appropriate unit of coordination is therefore the earliest conflicting step and the local reasoning that produced it.

Our approach, \method{}, centers deliberation on the earliest cross-agent conflict. Agents produce free-form solutions, which are converted into ordered nodes without altering the original answers. When the answers disagree, a localizer identifies the first conflict across these node sequences, and debate is limited to the short segments leading to that point. The controller checks the resolved claim against the question and the available evidence before adding it to the shared state. Agents then continue reasoning from this state and repeat the process if another conflict appears. We develop and evaluate \method{} on multi-hop question answering, where a single erroneous hop typically invalidates the final answer.

By treating cross-agent disagreement as a localized state-repair problem, our work makes three main contributions:
\begin{itemize}
    \item We formulate multi-agent debate as \emph{conflict-localized state repair}, resolving the earliest cross-agent divergence at each iteration while preserving committed state.
    \item We combine node extraction, indexed localization, local debate, guarded commitment, and early stopping in an inference-time protocol.
    \item On four multi-hop QA benchmarks and ten backbones, \method{} achieves the highest macro-averaged judge accuracy for every backbone, improving over the strongest conventional baseline by up to 7.20 percentage points.
\end{itemize}

\begin{figure*}[t]
\centering
\resizebox{\textwidth}{!}{%
\definecolor{lmadblue}{HTML}{0072B2}
\definecolor{lmadgreen}{HTML}{009E73}
\definecolor{lmadorange}{HTML}{D55E00}
\definecolor{lmadgold}{HTML}{E69F00}
\definecolor{lmadgray}{HTML}{7B8794}
\begin{tikzpicture}[
    x=1cm,y=1cm,>=Stealth,
    every node/.style={font=\scriptsize\sffamily,align=center},
    panel/.style={rounded corners=5pt,minimum height=3.05cm,inner sep=0pt,line width=0.75pt},
    heading/.style={font=\bfseries\scriptsize\sffamily,anchor=north},
    subtitle/.style={font=\tiny\sffamily,text=lmadgray,anchor=north},
    flow/.style={-{Stealth[length=4pt,width=5pt]},line width=0.9pt,draw=lmadgray},
    agent/.style={circle,draw=lmadblue,fill=lmadblue!12,minimum size=0.46cm,inner sep=0pt,font=\bfseries\tiny\sffamily},
    trace/.style={draw=lmadblue!65,fill=white,rounded corners=1.5pt,minimum width=0.50cm,minimum height=0.31cm,inner sep=1pt,font=\tiny\sffamily},
    verified/.style={draw=lmadgreen,fill=lmadgreen!14,rounded corners=1.5pt,minimum height=0.34cm,inner sep=1.5pt,font=\tiny\sffamily},
    conflict/.style={draw=lmadorange,line width=1.15pt,fill=lmadorange!16,rounded corners=1.5pt,minimum height=0.34cm,inner sep=1.5pt,font=\bfseries\tiny\sffamily},
    suffix/.style={draw=lmadgray,dashed,fill=lmadgray!7,rounded corners=1.5pt,minimum height=0.34cm,inner sep=1.5pt,font=\tiny\sffamily,text=lmadgray},
    segment/.style={draw=lmadorange,fill=lmadorange!10,rounded corners=2pt,minimum width=0.72cm,minimum height=0.38cm,inner sep=1pt,font=\bfseries\tiny\sffamily},
    check/.style={draw=lmadgold!85!black,fill=lmadgold!13,rounded corners=2pt,minimum width=2.12cm,minimum height=0.38cm,inner sep=1pt,font=\tiny\sffamily},
    state/.style={draw=lmadgreen,fill=lmadgreen!12,rounded corners=2pt,minimum height=0.38cm,inner sep=1.5pt,font=\tiny\sffamily},
    loop/.style={-{Stealth[length=4pt,width=5pt]},line width=0.85pt,dashed,draw=lmadblue}
]

\node[panel,draw=lmadblue!45,fill=lmadblue!4,minimum width=3.00cm] (p1) at (1.50,0) {};
\node[panel,draw=lmadgreen!50,fill=lmadgreen!4,minimum width=3.20cm] (p2) at (4.85,0) {};
\node[panel,draw=lmadorange!55,fill=lmadorange!4,minimum width=2.85cm] (p3) at (8.15,0) {};
\node[panel,draw=lmadgold!70!black,fill=lmadgold!5,minimum width=2.80cm] (p4) at (11.20,0) {};
\node[panel,draw=lmadgreen!50,fill=lmadgreen!4,minimum width=3.05cm] (p5) at (14.35,0) {};

\draw[flow] (p1.east) -- (p2.west);
\draw[flow] (p2.east) -- (p3.west);
\draw[flow] (p3.east) -- (p4.west);
\draw[flow] (p4.east) -- (p5.west);

\node[heading] at (1.50,1.40) {\textcolor{lmadblue}{\textcircled{1}} Parallel reasoning};
\node[subtitle] at (1.50,1.05) {independent chains};
\node[trace,minimum width=1.75cm] (query) at (1.50,0.52) {question $q$ + context};
\node[agent] (a1) at (0.45,0.02) {$A_1$};
\node[agent] (a2) at (0.45,-0.45) {$A_2$};
\node[agent] (a3) at (0.45,-0.92) {$A_3$};
\draw[draw=lmadblue!45,line width=0.5pt]
  (query.west) -- (0.13,0.52) -- (0.13,-0.92);
\foreach \y/\a/\i in {0.02/a1/1,-0.45/a2/2,-0.92/a3/3}{
  \node[trace] (zfirst\i) at (1.12,\y) {$z_1$};
  \node[trace] (zsecond\i) at (1.74,\y) {$z_2$};
  \node[trace] (zmore\i) at (2.36,\y) {$\cdots$};
  \draw[-{Stealth[length=2.6pt]},draw=lmadblue!60,line width=0.5pt]
    (0.13,\y) -- (\a.west);
  \draw[-{Stealth[length=2.6pt]},draw=lmadblue!70,line width=0.55pt]
    (\a.east) -- (zfirst\i.west);
  \draw[-{Stealth[length=2.6pt]},draw=lmadblue!70,line width=0.55pt]
    (zfirst\i.east) -- (zsecond\i.west);
  \draw[-{Stealth[length=2.6pt]},draw=lmadblue!70,line width=0.55pt]
    (zsecond\i.east) -- (zmore\i.west);
}

\node[heading] at (4.85,1.40) {\textcolor{lmadgreen}{\textcircled{2}} Align \& localize};
\node[subtitle] at (4.85,1.05) {locate earliest conflict};
\node[font=\tiny\sffamily,text=lmadgreen] at (4.02,0.52) {committed};
\node[font=\tiny\sffamily,text=lmadorange] at (4.93,0.52) {conflict};
\node[font=\tiny\sffamily,text=lmadgray] at (5.90,0.52) {untouched};
\foreach \y/\i in {0.12/1,-0.35/2,-0.82/3}{
  \node[font=\bfseries\tiny\sffamily,text=lmadgray] at (3.53,\y) {$A_{\i}$};
  \node[verified,minimum width=0.72cm] at (4.05,\y) {$P\;\checkmark$};
  \node[conflict,minimum width=0.82cm] (c\i) at (4.94,\y) {$z_{\i,t_{\i}}$};
  \node[suffix,minimum width=0.76cm] at (5.90,\y) {$\cdots$};
}
\draw[draw=lmadorange,rounded corners=3pt,line width=0.85pt] (4.43,0.36) rectangle (5.45,-1.05);
\node[fill=lmadorange,text=white,rounded corners=2pt,font=\bfseries\tiny\sffamily,inner sep=1.5pt] (localize) at (4.94,-1.20) {localize};

\node[heading] at (8.15,1.40) {\textcolor{lmadorange}{\textcircled{3}} Local debate};
\node[subtitle] at (8.15,1.05) {debate disputed spans};
\node[segment] (s1) at (7.35,0.38) {$S_1$};
\node[segment] (s2) at (7.35,-0.10) {$S_2$};
\node[segment] (s3) at (7.35,-0.58) {$S_3$};
\node[circle,draw=lmadorange,line width=1pt,fill=white,minimum size=0.95cm,font=\bfseries\tiny\sffamily] (debate) at (8.68,-0.10) {targeted\\debate};
\draw[-{Stealth[length=3pt]},draw=lmadorange,line width=0.7pt] (s1.east) -- (debate.west);
\draw[-{Stealth[length=3pt]},draw=lmadorange,line width=0.7pt] (s2.east) -- (debate.west);
\draw[-{Stealth[length=3pt]},draw=lmadorange,line width=0.7pt] (s3.east) -- (debate.west);
\node[suffix,text width=2.15cm] at (8.15,-1.08) {committed state + suffix stay fixed};

\node[heading] at (11.20,1.40) {\textcolor{lmadgold!80!black}{\textcircled{4}} Guarded commit};
\node[subtitle] at (11.20,1.05) {relation + evidence guards};
\node[conflict,minimum width=1.55cm] (proposal) at (11.20,0.42) {proposal $\widehat{Z}$};
\node[check] (relation) at (11.20,-0.08) {$\checkmark$ question relation};
\node[check] (evidence) at (11.20,-0.54) {$\checkmark$ evidence support};
\node[state,minimum width=1.55cm] (commit) at (11.20,-1.06) {commit $P\,\|\,\widehat{Z}$};
\draw[-{Stealth[length=3pt]},draw=lmadgold!80!black,line width=0.65pt]
  (proposal.south) -- (relation.north);
\draw[-{Stealth[length=3pt]},draw=lmadgold!80!black,line width=0.65pt]
  (relation.south) -- (evidence.north);
\draw[-{Stealth[length=3pt]},draw=lmadgold!80!black,line width=0.65pt]
  (evidence.south) -- (commit.north);

\node[heading] at (14.35,1.40) {\textcolor{lmadgreen}{\textcircled{5}} Continue or stop};
\node[subtitle] at (14.35,1.05) {repair, repeat, or answer};
\node[verified,minimum width=0.62cm] at (13.52,0.42) {$P$};
\node[state,minimum width=0.62cm] at (14.24,0.42) {$\widehat{Z}$};
\node[suffix,minimum width=0.72cm] at (15.05,0.42) {$\cdots$};
\node[trace,minimum width=1.82cm] (compare) at (14.35,-0.20) {compare updated answers};
\node[state,minimum width=1.62cm,font=\bfseries\tiny\sffamily] (answer) at (14.35,-0.96) {final answer};
\draw[-{Stealth[length=3pt]},draw=lmadgreen,line width=0.7pt]
  (compare.south) -- node[right,font=\tiny\sffamily,text=lmadgreen] {agree} (answer.north);

\draw[loop,rounded corners=4pt]
  (compare.west) -- (12.70,-0.20) -- (12.70,-1.72)
  -- node[midway,below,font=\bfseries\tiny\sffamily,text=lmadblue]
     {localize the next earliest conflict} (6.15,-1.72)
  .. controls (5.62,-1.72) and (5.18,-1.55) .. (localize.south);
\end{tikzpicture}
}
\caption{\textbf{System overview of \method{}.} Parallel rationales are conditioned on committed state $P$ and mapped to ordered chains. The first cross-agent conflict is localized, and only the segments $S_i$ ending at that conflict are debated. Relation and evidence guards admit a repair $\widehat{Z}$ into $P$. Agents resume from the updated state, localizing the next earliest conflict or stopping when answers agree. Green solid, orange solid, and gray dashed elements mark committed, disputed, and untouched states.}
\label{fig:overview}
\end{figure*}
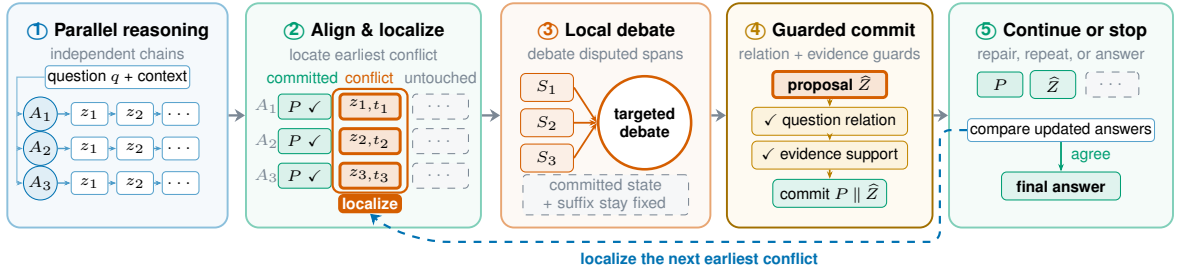

\section{Related Work}

\paragraph{Multi-agent debate.}
Multi-agent debate (MAD) coordinates multiple language-model agents through repeated proposal, critique, and revision. A canonical protocol exchanges complete solutions across rounds and aggregates the resulting answers, improving reasoning and factuality on several tasks~\citep{pmlr-v235-du24e}. Subsequent work has varied the composition of the debating group: ReConcile combines responses from diverse models through confidence-weighted consensus~\citep{chen-etal-2024-reconcile}, and divergent MAD assigns agents distinct perspectives to elicit competing lines of reasoning~\citep{liang-etal-2024-encouraging}. Recent studies target the exchange itself. Long debates tend to drift away from the original problem~\citep{becker-etal-2026-stay}, and revealed agent identities bias which arguments prevail~\citep{choi-etal-2026-identity}. SVR-MAD prunes agents and communication links under posterior guidance to cut debate cost~\citep{jiang-etal-2026-svrmad}, and Free-MAD replaces forced consensus with trajectory-level scoring and an anti-conformity mechanism~\citep{cui-etal-2026-free}.

Existing multi-agent debate methods primarily vary agent roles, interaction topology, reliability estimation, or answer aggregation. LMAD focuses on a different design choice: the unit of debate. It identifies the earliest conflict in the agents' rationales and restricts subsequent communication to the segments that lead to this conflict.

\section{Localized Multi-Agent Debate}
\label{sec:method}

Let $q$ be a question, $x$ its optional evidence context, and $A_1,\ldots,A_K$ homogeneous agents that share a base language model but use different sampling temperatures. The controller initializes an empty committed state $P^{(0)}$. At iteration $\ell$, agent $i$ samples a free-form rationale and answer conditioned on this state, $(r_i^{(\ell)},a_i^{(\ell)}) \sim A_i(\,\cdot\mid q,x,P^{(\ell)})$. \method{} then applies node extraction, conflict localization, local debate, and guarded commitment. Nodes in $P^{(\ell)}$ are not reopened in later iterations.

\subsection{Reasoning Nodes}

\method{} first elicits an ordinary step-by-step solution. An extractor $E$ maps the current solution to an ordered node sequence $Z_i^{(\ell)}=E(r_i^{(\ell)})=\bigl(z_{i,t}^{(\ell)}\bigr)_{t=1}^{T_i^{(\ell)}}$, where each node states one atomic claim. The extractor separately preserves $a_i^{(\ell)}$, preventing answer rewriting during structural parsing. Empty, repeated, excessively long, or malformed nodes are detected by a lightweight quality check.

\subsection{Indexed Conflict Localization}

If all normalized answers agree, the controller stops without debate. Otherwise, write the current agent states as $\mathcal{U}^{(\ell)}=\{(Z_i^{(\ell)},a_i^{(\ell)})\}_{i=1}^{K}$. A localizer $L$ returns a semantic conflict description $\delta^{(\ell)}$ and a one-based witness index for each agent:
\begin{equation}
 \bigl(\delta^{(\ell)},t_1^{(\ell)},\ldots,t_K^{(\ell)}\bigr)
 =L\bigl(q,x,P^{(\ell)},\mathcal{U}^{(\ell)}\bigr).
\end{equation}
The localizer scans the ordered chains from their beginnings and returns the first cross-agent inconsistency it encounters. For the common conflict $\delta^{(\ell)}$, $t_i^{(\ell)}$ indexes the first node in agent $i$'s chain that expresses or gives rise to that conflict. The localizer does not estimate downstream answer relevance. The debated segment is the prefix ending at this witness node, $S_i^{(\ell)}=\bigl(z_{i,t}^{(\ell)}\bigr)_{t=1}^{t_i^{(\ell)}}$, which carries the reasoning that produced the disputed claim; nodes after the witness build on that claim and are not debated but regenerated afresh from the updated state once the conflict is resolved. The agent-specific indices need not align numerically across chains. They make segment selection auditable and prevent fuzzy semantic matching from silently selecting a later disagreement.

\begin{table*}[!t]
\centering
\caption{Qwen2.5-32B judge accuracy (\%) across four benchmarks and ten
backbones. We compare \method{} with single-agent chain-of-thought
(CoT)~\citep{wei-etal-2022-chain}, three-sample self-consistency (SC)~\citep{wang-etal-2023-self-consistency}, iterative
Consensus (Cons.)~\citep{chen-etal-2024-reconcile}, standard multi-agent debate
(MAD)~\citep{pmlr-v235-du24e}, and divergent MAD
(dMAD)~\citep{liang-etal-2024-encouraging}. Predictions are evaluated by a
Qwen2.5-32B judge~\citep{yang-etal-2024-qwen25}. Underlined values indicate
the strongest conventional baseline in each row, and bold values indicate
the best overall result. The \textit{Avg.} rows report macro-averaged
accuracy across the four benchmarks. For these rows, $\Delta$ denotes the
improvement of \method{} over the strongest macro-averaged baseline, with
paired-bootstrap 95\% confidence intervals shown in brackets.}
\label{tab:main}
\footnotesize
\setlength{\tabcolsep}{6pt}

\begin{tabular}{llrrrrrrr}
\toprule
Backbone & Dataset
& \multicolumn{1}{c}{CoT}
& \multicolumn{1}{c}{SC}
& \multicolumn{1}{c}{Cons.}
& \multicolumn{1}{c}{MAD}
& \multicolumn{1}{c}{dMAD}
& \multicolumn{1}{c}{\method{}}
& \multicolumn{1}{c}{$\Delta$ [95\% CI]} \\
\midrule

Qwen2.5-7B
  & HotpotQA
  & 59.20 & 62.60 & 62.60 & \underline{63.00} & 55.00
  & \textbf{67.20} & +4.20 \\
  & 2Wiki
  & 66.40 & 68.80 & \underline{72.80} & 72.40 & 62.60
  & \textbf{76.00} & +3.20 \\
  & MuSiQue
  & 45.40 & 46.20 & \underline{47.80} & 47.00 & 34.20
  & \textbf{55.00} & +7.20 \\
  & StrategyQA
  & 89.60 & \underline{90.80} & 88.80 & 85.60 & 64.00
  & \textbf{94.00} & +3.20 \\
  & \textit{Avg.}
  & 65.15 & 67.10 & \underline{68.00} & 67.00 & 53.95
  & \textbf{73.05}
  & \textbf{+5.05} {\scriptsize[+3.50, +6.60]} \\

\cmidrule(lr){1-9}

Qwen2.5-14B
  & HotpotQA
  & 66.60 & \textbf{\underline{71.40}} & 68.00 & 68.20 & 64.40
  & 71.20 & $-0.20$ \\
  & 2Wiki
  & 78.20 & \underline{79.40} & 78.40 & 76.60 & 71.80
  & \textbf{81.20} & +1.80 \\
  & MuSiQue
  & 55.60 & 56.80 & \underline{58.40} & 56.80 & 48.20
  & \textbf{62.60} & +4.20 \\
  & StrategyQA
  & 87.80 & \underline{91.80} & 91.00 & 89.60 & 62.60
  & \textbf{93.40} & +1.60 \\
  & \textit{Avg.}
  & 72.05 & \underline{74.85} & 73.95 & 72.80 & 61.75
  & \textbf{77.10}
  & \textbf{+2.25} {\scriptsize[+1.00, +3.50]} \\

\cmidrule(lr){1-9}

Qwen2.5-32B
  & HotpotQA
  & 71.60 & \underline{73.00} & 71.40 & 70.20 & 65.40
  & \textbf{74.40} & +1.40 \\
  & 2Wiki
  & \underline{81.60} & 80.80 & \underline{81.60} & 80.80 & 77.20
  & \textbf{83.40} & +1.80 \\
  & MuSiQue
  & \underline{60.60} & 59.60 & 58.60 & 59.00 & 49.00
  & \textbf{62.60} & +2.00 \\
  & StrategyQA
  & \underline{94.20} & 94.00 & 92.20 & 90.00 & 76.60
  & \textbf{95.20} & +1.00 \\
  & \textit{Avg.}
  & \underline{77.00} & 76.85 & 75.95 & 75.00 & 67.05
  & \textbf{78.90}
  & \textbf{+1.90} {\scriptsize[+0.70, +3.15]} \\

\midrule

Qwen3-8B
  & HotpotQA
  & 79.00 & \textbf{\underline{80.80}} & 79.60 & 79.00 & 72.40
  & \textbf{80.80} & 0.00 \\
  & 2Wiki
  & 79.60 & 76.40 & \underline{80.20} & 78.60 & 73.20
  & \textbf{83.20} & +3.00 \\
  & MuSiQue
  & 54.60 & 55.80 & \underline{59.00} & 57.60 & 47.80
  & \textbf{63.40} & +4.40 \\
  & StrategyQA
  & \underline{91.80} & 91.40 & 90.80 & 88.00 & 68.40
  & \textbf{92.80} & +1.00 \\
  & \textit{Avg.}
  & 76.25 & 76.10 & \underline{77.40} & 75.80 & 65.45
  & \textbf{80.05}
  & \textbf{+2.65} {\scriptsize[+1.40, +3.90]} \\

\cmidrule(lr){1-9}

Qwen3-14B
  & HotpotQA
  & 82.80 & \underline{83.00} & 82.60 & 81.80 & 76.20
  & \textbf{84.60} & +1.60 \\
  & 2Wiki
  & 80.00 & 79.80 & \underline{81.20} & 80.40 & 77.00
  & \textbf{82.20} & +1.00 \\
  & MuSiQue
  & 57.00 & 56.60 & 59.00 & \underline{59.20} & 46.20
  & \textbf{63.40} & +4.20 \\
  & StrategyQA
  & \underline{93.00} & 92.00 & 89.40 & 87.00 & 68.20
  & \textbf{94.40} & +1.40 \\
  & \textit{Avg.}
  & \underline{78.20} & 77.85 & 78.05 & 77.10 & 66.90
  & \textbf{81.15}
  & \textbf{+2.95} {\scriptsize[+2.15, +4.50]} \\

\cmidrule(lr){1-9}

Qwen3-32B
  & HotpotQA
  & 80.80 & 82.60 & \underline{82.80} & 80.80 & 67.80
  & \textbf{83.40} & +0.60 \\
  & 2Wiki
  & 80.80 & \underline{81.40} & \underline{81.40} & 80.00 & 64.60
  & \textbf{82.40} & +1.00 \\
  & MuSiQue
  & 60.80 & 59.80 & \underline{61.80} & 59.20 & 38.20
  & \textbf{64.20} & +2.40 \\
  & StrategyQA
  & \underline{94.20} & 94.00 & 93.60 & 87.40 & 59.20
  & \textbf{94.80} & +0.60 \\
  & \textit{Avg.}
  & 79.15 & 79.45 & \underline{79.90} & 76.85 & 57.45
  & \textbf{81.20}
  & \textbf{+1.30} {\scriptsize[+0.15, +2.50]} \\

\midrule

Qwen3.5-4B
  & HotpotQA
  & 70.67 & 69.33 & \underline{72.67} & \underline{72.67} & 64.67
  & \textbf{79.33} & +6.66 \\
  & 2Wiki
  & 82.67 & 81.33 & \underline{83.33} & 81.33 & 75.33
  & \textbf{86.67} & +3.34 \\
  & MuSiQue
  & 57.33 & 54.67 & 65.33 & \underline{69.33} & 45.33
  & \textbf{76.00} & +6.67 \\
  & StrategyQA
  & 86.67 & \underline{90.00} & 89.33 & 76.67 & 70.00
  & \textbf{94.67} & +4.67 \\
  & \textit{Avg.}
  & 74.34 & 73.83 & \underline{77.67} & 75.00 & 63.83
  & \textbf{84.17}
  & \textbf{+6.50} {\scriptsize[+4.00, +9.00]} \\

\cmidrule(lr){1-9}

Qwen3.5-9B
  & HotpotQA
  & \underline{74.00} & \underline{74.00} & \underline{74.00}
  & 72.67 & 64.67 & \textbf{74.67} & +0.67 \\
  & 2Wiki
  & 79.33 & 81.33 & 81.33 & \underline{82.67} & 74.67
  & \textbf{84.67} & +2.00 \\
  & MuSiQue
  & 56.67 & 60.00 & \underline{66.67} & \underline{66.67} & 46.67
  & \textbf{71.33} & +4.66 \\
  & StrategyQA
  & 90.67 & 89.33 & \underline{91.33} & 87.33 & 68.67
  & \textbf{94.00} & +2.67 \\
  & \textit{Avg.}
  & 75.17 & 76.17 & \underline{78.33} & 77.34 & 63.67
  & \textbf{81.17}
  & \textbf{+2.84} {\scriptsize[+0.67, +5.00]} \\

\midrule

Gemma3-4B
  & HotpotQA
  & 68.20 & 69.40 & \underline{72.60} & 72.40 & 53.20
  & \textbf{75.00} & +2.40 \\
  & 2Wiki
  & 61.20 & 62.00 & 67.40 & \textbf{\underline{68.00}} & 49.20
  & 64.20 & $-3.80$ \\
  & MuSiQue
  & 32.00 & 35.60 & 41.20 & \underline{42.80} & 28.40
  & \textbf{44.00} & +1.20 \\
  & StrategyQA
  & \underline{91.40} & 91.20 & 87.20 & 83.80 & 43.80
  & \textbf{92.20} & +0.80 \\
  & \textit{Avg.}
  & 63.20 & 64.55 & \underline{67.10} & 66.75 & 43.65
  & \textbf{68.85}
  & \textbf{+1.75} {\scriptsize[+0.15, +3.40]} \\

\cmidrule(lr){1-9}

Gemma3-12B
  & HotpotQA
  & \underline{79.00} & 78.00 & 78.40 & 78.00 & 61.60
  & \textbf{81.60} & +2.60 \\
  & 2Wiki
  & 78.20 & 77.40 & \textbf{\underline{80.00}} & 79.40 & 62.20
  & 79.80 & $-0.20$ \\
  & MuSiQue
  & 54.00 & 56.20 & \underline{57.00} & 56.40 & 37.80
  & \textbf{60.60} & +3.60 \\
  & StrategyQA
  & \underline{95.80} & 95.60 & 94.80 & 91.40 & 39.80
  & \textbf{97.00} & +1.20 \\
  & \textit{Avg.}
  & 76.75 & 76.80 & \underline{77.55} & 76.30 & 50.35
  & \textbf{79.75}
  & \textbf{+2.20} {\scriptsize[+0.90, +3.50]} \\

\bottomrule
\end{tabular}
\end{table*}

\subsection{Local Debate}

Each agent receives $(q,x)$, the committed state $P^{(\ell)}$, its own localized segment $S_i^{(\ell)}$, and brief descriptions of the competing claims. Positions are initialized to the witness claims, $p_i^{(0)}=z_{i,t_i^{(\ell)}}^{(\ell)}$. In each of $R$ local rounds, every agent defends or revises its position in view of the others',
\begin{equation}
p_i^{(\rho)}
\sim A_i\!\bigl(\,\cdot\mid q,x,P^{(\ell)},S_i^{(\ell)},\delta^{(\ell)},
\{p_j^{(\rho-1)}\}_{j\neq i}\bigr).
\end{equation}
A zero-temperature resolver $\Phi$ then turns the final positions into a proposed repair,
\begin{equation}
\bigl(\widehat Z^{(\ell)},\widehat Q^{(\ell)}\bigr)
=\Phi\bigl(q,x,P^{(\ell)},\delta^{(\ell)},
\{p_i^{(R)}\}_{i=1}^{K}\bigr),
\end{equation}
where $\widehat Z^{(\ell)}$ contains one or more replacement nodes and $\widehat Q^{(\ell)}$ their supporting quotations. The commit guard admits the repair only if the question's subject and relation are preserved, and every quotation appears verbatim in the provided context. The committed state is updated as
\begin{equation}
P^{(\ell+1)}
=
\begin{cases}
P^{(\ell)}\mathbin{\|}\widehat Z^{(\ell)}, & \text{if admitted},\\
P^{(\ell)}, & \text{otherwise}.
\end{cases}
\end{equation}
On commitment, the agents begin another iteration conditioned on $P^{(\ell+1)}$; otherwise the controller falls back to a majority vote, also used when the iteration budget is exhausted. The process terminates when the normalized answers agree, a proposed repair fails the commit guard, or the maximum number of iterations is reached.

\section{Experiments}
Our experiments evaluate whether \method{} yields consistent improvements across model families, parameter scales, and datasets (see Appendix~\ref{app:experimental-details} for experimental details).

\paragraph{Datasets and models.}
Our evaluation spans HotpotQA~\citep{yang-etal-2018-hotpotqa}, 2WikiMultiHopQA~\citep{ho-etal-2020-constructing}, MuSiQue~\citep{Trivedi_2022}, and StrategyQA~\citep{Geva_2021}. The experiments use ten open-weight backbones from four model families: Qwen2.5~\citep{yang-etal-2024-qwen25}, Qwen3~\citep{yang-etal-2025-qwen3}, Qwen3.5~\citep{qwen-team-2026-qwen35}, and Gemma3~\citep{gemma-team-etal-2025-gemma3}.

\paragraph{Main results.}

Table~\ref{tab:main} shows that \method{} achieves the highest macro-averaged judge accuracy for all ten backbones, improving over the strongest macro-averaged baseline by 1.30 to 6.50 percentage points. Across the 40 backbone--dataset combinations, \method{} improves over the strongest baseline in 36 cases, ties in one, and underperforms in three. The gains are particularly consistent on MuSiQue, where \method{} outperforms the strongest baseline for every backbone and reaches a maximum improvement of 7.20 points. Since MuSiQue contains reasoning chains of up to four hops, this pattern suggests that localizing the earliest divergence is especially helpful when errors can propagate through longer chains. Overall, these results show that localized debate transfers across model families, parameter scales, and reasoning depths. Appendix~\ref{app:qualitative} provides qualitative examples.

\paragraph{Commitment-gate ablation.}
To examine whether guarded commitment is necessary, we construct a gate-activated subset containing questions for which the default strict policy rejected at least one candidate repair.
We compare three commitment policies on the same questions under otherwise identical settings.
\textsc{Strict}, the default \method{} policy, requires the proposed repair to preserve the relevant relation and be supported by the available evidence. 
\textsc{Concrete} only requires the resolver output to form an explicit claim, without checking relation preservation or evidential support. 
\textsc{Always} commits every resolver proposal that reaches the gate.
Table~\ref{tab:gate_ablation} shows \textsc{Strict} achieves a macro-average accuracy of 55.98\%, outperforming \textsc{Concrete} and \textsc{Always} by 6.56 and 7.56 percentage points, respectively. 

\begin{table}[!h]
\centering
\footnotesize
\setlength{\tabcolsep}{3.2pt}
\caption{
Accuracy (\%) on gate-activated subsets containing instances
for which \textsc{Strict} rejected at least one candidate repair.
}
\label{tab:gate_ablation}
\begin{tabular}{lrrr}
\toprule
Dataset & \textsc{Strict} & \textsc{Concrete} & \textsc{Always} \\
\midrule
HotpotQA    & \textbf{59.83} & 50.43 & 48.72 \\
2Wiki       & \textbf{53.95} & 51.24 & 50.92 \\
MuSiQue     & \textbf{36.61} & 34.25 & 32.28 \\
StrategyQA  & \textbf{73.53} & 61.76 & 61.76 \\
\midrule
Macro Avg.  & \textbf{55.98} & 49.42 & 48.42 \\
\bottomrule
\end{tabular}
\end{table}

\section{Conclusion}

We introduced \method{}, a multi-agent debate framework that localizes the earliest unresolved cross-agent divergence, debates only the reasoning steps leading to it, and commits resolved claims to a shared state. Across four multi-hop QA benchmarks and ten backbones from four model families, \method{} improves macro-averaged judge accuracy over the strongest conventional baseline by 1.30 to 6.50 percentage points, with an average gain of 2.94 points. These results support conflict localization as an effective principle for multi-agent debate.

\clearpage
\section*{Limitations}
Our current formulation represents reasoning as an ordered chain of intermediate states, whereas real-world reasoning may exhibit richer, non-linear structures. Future work could extend localized debate to tree- or graph-structured representations, including directed acyclic graphs, to capture branching hypotheses, converging evidence, and non-linear dependencies.

\bibliography{custom}

\clearpage

\appendix
\section{Additional Experimental Details}
\label{app:experimental-details}

\subsection{Datasets}

\paragraph{HotpotQA.}
HotpotQA is an explainable multi-hop QA benchmark built from Wikipedia
articles~\citep{yang-etal-2018-hotpotqa}. Its two principal reasoning patterns
are \emph{bridge} questions, where an intermediate entity found in one article
identifies the second article needed for the answer, and \emph{comparison}
questions, where properties of two entities must be aligned and compared. In
the distractor setting used here, the supporting articles are mixed with
semantically related but non-supporting articles, yielding ten candidate
paragraphs per example. The benchmark annotates the supporting sentences, which
allows the evidence chain to be audited in addition to the usually short-span
or yes/no answer.

\paragraph{2WikiMultiHopQA.}
2WikiMultiHopQA combines Wikipedia passages with relations derived from
Wikidata and was designed to expose intermediate reasoning steps~\citep{ho-etal-2020-constructing}. It covers
compositional, comparison, inference, and bridge-comparison questions. These
categories require, respectively, chaining relations, comparing two entities,
inferring an unstated relation from available facts, or first resolving a
bridge entity and then performing a comparison. Examples provide supporting
evidence alongside distractor passages, while answers are short entities,
attribute values, or binary decisions.

\paragraph{MuSiQue.}
MuSiQue constructs multi-hop questions by composing independently answerable
single-hop questions into connected chains~\citep{Trivedi_2022}. Its
construction procedure is intended to reduce shortcut solutions in which a
model can ignore part of the chain and still recover the answer. The benchmark
contains two-, three-, and four-hop instances, mixes the supporting paragraphs
with distractors, and supplies aliases for valid short answers. It therefore
tests whether a system can maintain the correct intermediate entity across a
longer evidence chain.

\paragraph{StrategyQA.}
StrategyQA contains binary questions whose required reasoning strategy is not
stated explicitly in the question~\citep{Geva_2021}. A solver must first infer
an appropriate decomposition into factual subquestions and then combine the
resulting facts to produce \emph{yes} or \emph{no}. This differs from the three
Wikipedia-centered benchmarks: the main difficulty is deciding which implicit
facts and relations are needed. The original benchmark includes evidence
facts and decomposition annotations; our prompts include the supplied facts
but do not reveal the annotated decomposition.

All runs use the full context supplied with each example. For HotpotQA,
2WikiMultiHopQA, and MuSiQue, this includes both supporting passages and the
provided distractor passages; for StrategyQA, it includes the supplied facts.

\subsection{Evaluation Protocol}

The primary metric in Table~\ref{tab:main} is judge accuracy. A
Qwen2.5-32B evaluator compares each saved prediction with the reference answer
and, under deterministic decoding, returns a binary semantic-correctness
decision. Judge accuracy is the percentage of correct decisions. The same
judge protocol is applied to all methods, and the reported macro-average gives
equal weight to the four datasets.

\subsection{Baseline and Inference Configuration}

We use three agents for every multi-agent method and keep the backbone fixed
within each comparison. The conventional baselines are configured as follows.
\begin{itemize}
    \item \textbf{CoT} generates one chain-of-thought solution with temperature
    $0$.
    \item \textbf{SC} generates three solutions and
    returns a majority vote after answer normalization.
    \item \textbf{MAD} performs an independent initial generation for each of
    three agents followed by two answer-level revision rounds. This gives nine
    solver calls per example, after which normalized majority voting determines
    the prediction.
    \item \textbf{Consensus} uses the same three-agent exchange and stops early
    when all normalized answers agree; otherwise, it returns a majority vote
    after at most two revision rounds.
    \item \textbf{dMAD} assigns opposing affirmative and negative roles. A judge checks the debate after each round and stops when it
    can decide; if no decision is reached, a moderator returns
    the final answer.
\end{itemize}

\subsection{Implementation details}
All experiments are conducted with vLLM using four NVIDIA A100 80GB GPUs. The total computational cost is approximately 60 GPU-hours. For
each evaluated backbone, the same model is used for rationale generation and
all auxiliary modules, including node extraction, conflict localization, and
resolution. The three solver personas use sampling temperatures
$(0.1,0.5,0.9)$, while all auxiliary modules use deterministic decoding with
temperature $0$. The reference controller performs one local debate round per
iteration and runs for at most ten iterations. Qwen3 and Qwen3.5 are run with
their explicit thinking mode disabled.

\section{Qualitative Examples}
\label{app:qualitative}

This appendix follows \method{} end to end on two evaluation
instances, from the question and evidence context, through the
extracted chains, localization, local debate, and guarded
commitment, to the final answer. Both traces use the Qwen3-8B backbone
with the fixed configuration of
Appendix~\ref{app:experimental-details}. In both instances the first
parallel pass yields a two-to-one majority for a wrong answer, so
majority voting over the initial answers fails, and the runs
illustrate how \method{} instead repairs the divergence at its source.
Figures~\ref{fig:case1} and~\ref{fig:case2} reproduce the traces. Node
claims, debate turns, and quotations are verbatim from the recorded
traces; bracketed ellipses mark abridged text, and the evidence
context is condensed to its decisive sentences.

\subsection{Case 1: Repairing an Off-by-One Lineage Error
(2WikiMultiHopQA)}
\label{app:case1}

The question asks for the paternal grandfather of Arthur Gore, 2nd
Earl of Arran. As Figure~\ref{fig:case1} shows, all three agents
recover the same two parent facts, from the 2nd Earl to Arthur Gore,
1st Earl of Arran, and from the 1st Earl to Sir Arthur Gore, 2nd
Baronet. Agents $A_1$ and $A_3$ nevertheless continue up the lineage
through Paul Gore and answer with the more distant Sir Arthur Gore,
1st Baronet, an off-by-one generation error; only $A_2$ stops at the
second hop. The initial vote is therefore two to one for the wrong
entity, and on this instance CoT, SC, Consensus, and MAD all return
the 1st Baronet as well. The localizer confines the conflict to the
final deduction nodes, since the antecedent parent facts agree across
chains, and in a single local round both erring agents re-derive the
two-hop lineage and drop the extra hop. The resolver's replacement
node passes the relation and quotation checks against the cited
passages, and the second iteration is unanimously correct.

\begin{figure*}[t]
\begin{casetrace}{Case 1 $\cdot$ 2WikiMultiHopQA $\cdot$ Qwen3-8B
$\cdot$ instance \texttt{27e835ee0bb011ebab90acde48001122}}

\noindent\textbf{Question.} Who is Arthur Gore, 2nd Earl of Arran's
paternal grandfather?\\
\textbf{Gold answer.} \emph{Sir Arthur Gore, 2nd Baronet}\\[2pt]
\textbf{Key evidence in the supplied context} (the passages later
cited by the resolver):
\begin{itemize}[label={},leftmargin=1.2em,itemsep=1pt,topsep=1pt]
\item ``Arthur Gore, 1st Earl of Arran PC (Ire) (1703 – 17 April
1773), known as Sir Arthur Gore, 3rd Baronet from 1741 to 1757 and as
Viscount Sudley from 1758 to 1762, was an Irish politician. Arran was
the son of Sir Arthur Gore, 2nd Baronet, and Elizabeth Annesley''
\item ``Sir Arthur Gore, 2nd Baronet( c. 1685 – 10 February 1742) was
an Irish politician and baronet. He was the son of Paul Gore, himself
son of Sir Arthur Gore, 1st Baronet, and his wife Anne Gore, daughter
of Sir John Gore.''
\end{itemize}

\casestage{lmadblue}{\textcircled{1}\; First iteration: independent chains}

\noindent $A_1$, answer \emph{Sir Arthur Gore, 1st Baronet}~\errmark
\begin{itemize}[label={},leftmargin=1.6em,itemsep=0.5pt,topsep=1pt]
\item $z_{1,1}$ \nodetag{fact} Arthur Gore, 2nd Earl of Arran was the
eldest son of Arthur Gore, 1st Earl of Arran and Jane Saunders.
\item $z_{1,2}$ \nodetag{fact} Arthur Gore, 1st Earl of Arran was the
son of Sir Arthur Gore, 2nd Baronet and Elizabeth Annesley.
\item $z_{1,3}$ \nodetag{fact} Sir Arthur Gore, 2nd Baronet was the
son of Paul Gore, who was the son of Sir Arthur Gore, 1st Baronet and
Anne Gore.
\item $z_{1,4}$ \nodetag{deduction} \textcolor{lmadorange}{Therefore,
the paternal grandfather of Arthur Gore, 2nd Earl of Arran is Sir
Arthur Gore, 1st Baronet.}
\end{itemize}
\smallskip
\noindent $A_2$, answer \emph{Sir Arthur Gore, 2nd Baronet}~\okmark
\begin{itemize}[label={},leftmargin=1.6em,itemsep=0.5pt,topsep=1pt]
\item $z_{2,1}$, $z_{2,2}$ \nodetag{fact} verbatim identical to
$z_{1,1}$ and $z_{1,2}$.
\item $z_{2,3}$ \nodetag{deduction} \textcolor{lmadorange}{Sir Arthur
Gore, 2nd Baronet is the paternal grandfather of Arthur Gore, 2nd Earl
of Arran.}
\end{itemize}
\smallskip
\noindent $A_3$, answer \emph{Sir Arthur Gore, 1st Baronet}~\errmark
\begin{itemize}[label={},leftmargin=1.6em,itemsep=0.5pt,topsep=1pt]
\item $z_{3,1}$ \nodetag{fact} Arthur Gore, 2nd Earl of Arran's father
is Arthur Gore, 1st Earl of Arran.
\item $z_{3,2}$ \nodetag{fact} Arthur Gore, 1st Earl of Arran's father
is Sir Arthur Gore, 2nd Baronet.
\item $z_{3,3}$ \nodetag{fact} Sir Arthur Gore, 2nd Baronet's father
is Paul Gore.
\item $z_{3,4}$ \nodetag{fact} Paul Gore's father is Sir Arthur Gore,
1st Baronet.
\item $z_{3,5}$ \nodetag{deduction} \textcolor{lmadorange}{Sir Arthur
Gore, 1st Baronet is the paternal grandfather of Arthur Gore, 2nd Earl
of Arran.}
\end{itemize}

\casestage{lmadgreen}{\textcircled{2}\; Align and localize}
\noindent The answers disagree, so the localizer scans the chains and
reports the earliest cross-agent conflict, ``Conflict in the
identification of the paternal grandfather of Arthur Gore, 2nd Earl of
Arran,'' with witness nodes $t_1{=}4$, $t_2{=}3$, $t_3{=}5$ (orange
above). The disagreement is confined to the final deduction of each
chain; the parent facts agree across all three chains.

\casestage{lmadorange}{\textcircled{3}\; Local debate (one round)}
\noindent Each agent receives its own segment, the conflict
description, and the competing claims, then defends or revises its
position:
\begin{itemize}[label={},leftmargin=1.6em,itemsep=0.5pt,topsep=1pt]
\item $A_1$: ``Arthur Gore, 2nd Earl of Arran's father was Arthur
Gore, 1st Earl of Arran. Arthur Gore, 1st Earl of Arran's father was
Sir Arthur Gore, 2nd Baronet. Therefore, Sir Arthur Gore, 2nd Baronet
is the paternal grandfather of Arthur Gore, 2nd Earl of Arran.
[\dots]''
\item $A_2$: ``Arthur Gore, 2nd Earl of Arran's father was Arthur
Gore, 1st Earl of Arran, who was the son of Sir Arthur Gore, 2nd
Baronet and Elizabeth Annesley. Therefore, Sir Arthur Gore, 2nd
Baronet is the paternal grandfather [\dots]''
\item $A_3$: ``Re-examining the segment: [\dots] Therefore, Sir
Arthur Gore, 2nd Baronet is the paternal grandfather of Arthur Gore,
2nd Earl of Arran.''
\end{itemize}

\casestage{lmadgold!80!black}{\textcircled{4}\; Guarded commit}
\noindent Resolver proposal $\widehat{Z}$:\\[2pt]
\commitclaim{Arthur Gore, 2nd Earl of Arran's paternal grandfather is
Sir Arthur Gore, 2nd Baronet}\\[2pt]
\noindent The proposal cites the two passages quoted at the top of
this box. The commit guard confirms that the question's subject and
relation are preserved and that both quotations appear verbatim in the
context, then appends the node to the committed state $P$.

\casestage{lmadgreen}{\textcircled{5}\; Second iteration and stop}
\noindent Conditioned on $P$, all three agents regenerate their chains
and answer \emph{Sir Arthur Gore, 2nd
Baronet}~\okmark\okmark\okmark. The normalized answers agree, and the
controller stops. $A_3$'s regenerated chain still lists the deeper
ancestry (Paul Gore, then the 1st Baronet) but now halts the deduction
at the correct generation.

\end{casetrace}
\caption{\textbf{Case 1 trace.} \method{} on a 2WikiMultiHopQA
instance with the Qwen3-8B backbone. Two of three agents traverse one
generation too far, so the initial vote favors the wrong entity two to
one; CoT, SC, Consensus, and MAD also fail on this instance.
Localization isolates the final deduction node of each chain (orange),
one local debate round restores the two-hop lineage, and the guarded
commit (green) repairs the shared state, after which the second
iteration is unanimously correct. Node claims, debate turns, and
quotations are verbatim from the recorded trace; bracketed ellipses
mark abridgment.}
\label{fig:case1}
\end{figure*}

\subsection{Case 2: Repairing an Answer-Granularity Error (HotpotQA)}
\label{app:case2}

The question introduces Hayden as a Canadian singer-songwriter and
asks where the band Buck-Tick hail from; the gold answer is the band's
formation place, Fujioka, Gunma. The decisive context sentence
supports two readings at different granularities, a nationality (``a
Japanese rock band'') and a formation place (``formed in Fujioka,
Gunma in 1983''). As Figure~\ref{fig:case2} shows, $A_1$ selects the
formation place, while $A_2$ and $A_3$ answer with the country.
Notably, $A_3$'s chain contains the correct formation fact, so its
error lies in the granularity of the final answer node rather than in
evidence selection. The localizer flags the earliest material claim
about Buck-Tick's origin in each chain, and a single debate round
re-anchors all three agents on the same sentence, after which both
country-answering agents adopt the specific location. The committed
node yields a unanimous second iteration.

\begin{figure*}[t]
\begin{casetrace}{Case 2 $\cdot$ HotpotQA $\cdot$ Qwen3-8B $\cdot$
instance \texttt{5ae4a3265542995ad6573de5}}

\noindent\textbf{Question.} Hayden is a singer-songwriter from Canada;
where do Buck-Tick hail from?\\
\textbf{Gold answer.} \emph{Fujioka, Gunma}\\[2pt]
\textbf{Key evidence in the supplied context} (the sentences quoted by
the agents):
\begin{itemize}[label={},leftmargin=1.2em,itemsep=1pt,topsep=1pt]
\item ``Paul Hayden Desser (born February 12, 1971) who records as
Hayden, is a Canadian singer-songwriter from Thornhill, Ontario.''
\item ``Buck-Tick (stylized as BUCK-TICK) is a Japanese rock band,
formed in Fujioka, Gunma in 1983.''
\end{itemize}

\casestage{lmadblue}{\textcircled{1}\; First iteration: independent chains}

\noindent $A_1$, answer \emph{Fujioka, Gunma}~\okmark
\begin{itemize}[label={},leftmargin=1.6em,itemsep=0.5pt,topsep=1pt]
\item $z_{1,1}$ \nodetag{fact} \textcolor{lmadorange}{Buck-Tick was
formed in Fujioka, Gunma.}
\item $z_{1,2}$ \nodetag{answer} Buck-Tick hail from Fujioka, Gunma.
\end{itemize}
\smallskip
\noindent $A_2$, answer \emph{Japan}~\errmark
\begin{itemize}[label={},leftmargin=1.6em,itemsep=0.5pt,topsep=1pt]
\item $z_{2,1}$ \nodetag{fact} \textcolor{lmadorange}{Hayden is a
Canadian singer-songwriter.}
\item $z_{2,2}$ \nodetag{fact} Buck-Tick is a Japanese rock band.
\item $z_{2,3}$ \nodetag{answer} Buck-Tick hail from Japan.
\end{itemize}
\smallskip
\noindent $A_3$, answer \emph{Japan}~\errmark
\begin{itemize}[label={},leftmargin=1.6em,itemsep=0.5pt,topsep=1pt]
\item $z_{3,1}$ \nodetag{fact} \textcolor{lmadorange}{Buck-Tick is a
Japanese rock band.}
\item $z_{3,2}$ \nodetag{fact} Buck-Tick was formed in Fujioka, Gunma.
\item $z_{3,3}$ \nodetag{answer} Buck-Tick hails from Japan.
\end{itemize}
{\color{lmadgray}\itshape $A_3$ extracts the correct formation fact
($z_{3,2}$) yet still answers with the country.}

\casestage{lmadgreen}{\textcircled{2}\; Align and localize}
\noindent The localizer reports ``Conflict in the earliest material
claim about the origin of Buck-Tick,'' with witness nodes
$t_1{=}t_2{=}t_3{=}1$ (orange above), the earliest node in each chain
that gives rise to the conflicting origin claims.

\casestage{lmadorange}{\textcircled{3}\; Local debate (one round)}
\begin{itemize}[label={},leftmargin=1.6em,itemsep=0.5pt,topsep=1pt]
\item $A_1$: ``The passage states that Buck-Tick is a Japanese rock
band formed in Fujioka, Gunma. This directly answers the question
about the origin of Buck-Tick. There is no conflicting information
regarding the band's origin in the provided text. [\dots]''
\item $A_2$: ``The passage about Buck-Tick states: `Buck-Tick
(stylized as BUCK-TICK) is a Japanese rock band, formed in Fujioka,
Gunma in 1983.' This directly confirms that Buck-Tick hail from
Fujioka, Gunma.''
\item $A_3$: ``Buck-Tick is explicitly stated as a Japanese rock band
formed in Fujioka, Gunma. [\dots] My claim: Buck-Tick hail from
Fujioka, Gunma.''
\end{itemize}

\casestage{lmadgold!80!black}{\textcircled{4}\; Guarded commit}
\noindent Resolver proposal $\widehat{Z}$:\\[2pt]
\commitclaim{Buck-Tick hail from Fujioka, Gunma.}\\[2pt]
\noindent The supporting quotation is the Buck-Tick sentence quoted at
the top of this box. The commit guard verifies relation preservation
and the verbatim quotation, then appends the node to the committed
state $P$.

\casestage{lmadgreen}{\textcircled{5}\; Second iteration and stop}
\noindent Conditioned on $P$, all three agents answer \emph{Fujioka,
Gunma}~\okmark\okmark\okmark, and the controller stops on answer
agreement.

\end{casetrace}
\caption{\textbf{Case 2 trace.} \method{} on a HotpotQA instance with
the Qwen3-8B backbone. The decisive sentence supports both a
country-level and a city-level reading of the band's origin; two
agents initially answer \emph{Japan}, one of them despite extracting
the formation fact. Localization flags the earliest origin claims
(orange), a single debate round aligns all agents on the formation
place, and the committed repair (green) produces a unanimous second
iteration. Node claims, debate turns, and quotations are verbatim from
the recorded trace; bracketed ellipses mark abridgment.}
\label{fig:case2}
\end{figure*}

\subsection{Why the Repairs Succeed}
\label{app:why-repairs}

Both traces begin in a state that answer-level aggregation cannot
repair, since the correct answer is a one-vote minority. Three
properties of \method{} account for the recoveries. First, the actual
disagreement is much narrower than the answer-level disagreement
suggests. In Case 1 the agents share every antecedent fact and diverge
only in the final deduction; in Case 2 one country-answering agent
already holds the decisive formation fact. Localization makes this
structure explicit and keeps the shared material out of dispute.
Second, focused verification is more reliable than free-form
generation. Once the comparison is reduced to a short segment, an
explicit conflict description, and competing claims, the agents that
produced the error identify the correct claim within a single round in
both traces, overturning wrong two-to-one majorities rather than
ratifying them. Third, guarded commitment converts local agreement
into shared state. A repair is admitted only after the relation and
verbatim-quotation checks pass, and every agent reconditions on the
committed node, which yields unanimous second iterations and early
termination in both cases.

\end{document}